\documentclass[10pt,towcolumn,journal]{IEEEtran}
\usepackage{graphicx,amssymb,amsmath,amsthm,cite,cases}
\usepackage{adjustbox}

\usepackage{color,multirow}
\usepackage{mathrsfs}
\DeclareGraphicsExtensions{.tif}

\def\oa{\hat{\operatorname{J}}}

\def\be{\begin{equation}}
\def\ee{\end{equation}}
\def\ben{\begin{eqnarray}}
\def\een{\end{eqnarray}}

\def\D{\mathcal{D}}
\def\R{\mathbb{R}}

\def\V{\mathbb{V}}

\def\kqd{k_{\qd}}

\def\jqd{j^{\qd}}
\def\qd{{q^\diamond}}

\def\til{\tilde}
\def\Nqq{\Nq \times \Nq}

\def\vI{\mathbf{I}}
\def\vG{\mathbf{G}}

\def\vR{\mathbf{R}}
\def\vd{\mathbf{d}}
\def\vD{\mathbf{D}}
\def\vc{\mathbf{c}}

\def\vA{\mathbf{A}}
\def\vB{\mathbf{B}}
\def\vT{\mathbf{T}}
\def\vW{\mathbf{U}}
\def\vU{\mathbf{W}}
\def\vRe{\mathbf{R}}
\def\vR{\mathbf{R}}
\def\vRt{\til{\mathbf{R}}}

\def\vd{\mathbf{d}}

\def\op{\hat{P}}
\def\kq{k_q}
\def\Nq{N_b}

\def\PSNR{\mathrm{PSNR}}
\def\SR{\mathrm{SR}}

\def\G{\mathrm{G}}
\def\MSE{\mathrm{MSE}}

\def\Mx{M_x}
\def\My{M_y}

\newcommand{\la}{\left \langle}
\newcommand{\ra}{\right \rangle}

\def\Nx{N_x}
\def\Ny{N_y}
\def\Nxy{\Nx \times \Ny}

\def\Mx{M_x}
\def\My{M_y}

\newcommand{\Spann}{{\mbox{\rm{span}}}}

\title{Effective sparse representation of
 X-Ray medical images}
\author{Laura Rebollo-Neira\\
Mathematics Department\\
Aston University\\
B4 7ET Birmingham, UK}
\date{}
\begin{document}
\maketitle
\begin{abstract}
Effective sparse representation of X-Ray medical images 
within the context of data reduction is considered.  
The proposed framework is shown to render an enormous 
reduction in the cardinality of the data set 
required to represent this class of images  
at very good quality. The particularity of the 
approach is that it can be implemented at very 
competitive processing time and low memory 
requirements.\\\\
\end{abstract}
\section{Introduction}
Within the field of medical imaging for diagnosis, 
radiology generates huge volumes of data in the 
form of X-Ray images. 
Complying with archive provisions legislation, which may 
require to store the 
patient's data for up to ten years, represents a demanding 
 burden for hospitals and individual radiology practices.
Additionally, the prompt distribution of remote radiology 
reporting is one of the challenges in teleradiology. 
 These matters have led 
 several radiological societies to recommend to 
use  irreversible (or `lossy') compression  
 ``in a manner that 
is visually imperceptible and/or without loss of diagnostic 
performance'' \cite{ESR11}.

At least for extensive use, the state of the art 
for lossy image compression are the 
  JPEG and JPEG2000 standards.
 Both techniques belong to 
the category of transformation coding, because are based 
on a compression scheme that applies an invertible  
transformation as 
the first step in the process. JPEG uses the 
Discrete Cosine Transform (DCT) for that purpose and 
JPEG2000 the Discrete Wavelet Transform (DWT). 
Both transformations play the role of reducing 
the non-negligible points in the transformed domain.  
The transformation we adopt here for the same purpose 
is different in essence. Rather than transforming the 
data into an array of the same dimensionality to  
disregard some points there, 
we expand the representation domain and strive to
achieve a highly sparse representation in the extended
domain.

Apart from the perceived advantage of sparse 
representations for medical image  processing  and 
health informatics \cite{APM15}, 
the emerging theory of compressive sensing 
has introduced a strong reason to achieve sparsity.
Within the compressive sensing structure the number of
measurements needed for accurate representation of
a signal informational content
decreases if the sparsity of the representation improves
\cite{Don06,CW08,Bar11}.

This Communication presents a framework rendering 
 high sparsity in the representation of X-Ray medical 
 images. This is achieved by: 
\begin{itemize}
\item[(a)] Creating a large redundant `dictionary' 
of suitable elements for the image decomposition. 
\item[(b)] Applying effective strategies for 
selecting the particular elements
 which enable the sparse decomposition of a given
image.
\end{itemize}
The goal is to achieve high sparsity, with high quality
reconstruction, at competitive processing time. 
Comparison of the results arising from the proposed 
framework with those yielded by the traditional 
 DCT or DWT approximations demonstrates a massive 
improvement in sparsity.
\section{Sparse Image Representation}
Let's start by introducing some notational convention: 
bold face lower and upper cases are used to represent 
one dimension (1D) and two dimension (2D) arrays, 
respectively. Standard mathematical fonts indicate
component, e.g., $\vc \in \R^K$ is an array of 
real components, $c(k),\, k=1,\ldots,K$, and 
$\vI \in \R^{\Nxy}$ an array of real elements,
$I(i,j),\,i=1,\ldots,\Nx,\,i=1,\ldots,\Ny$.

Restricting considerations to $l$-bit gray scale images, 
 an image is represented by an array 
$\vI \in \R^{\Nxy}$  the elements of which, 
called intensity pixels, 
are given by integer numbers from 0 to $2^l$-$1$.  

Within the adopted framework for representations  
using dictionaries an image 
$\vI \in \R^{\Nxy}$ is approximated by a linear
decomposition of the form:
\be
\vI^K= \sum_{k=1}^K c(k) \vD_{\ell_k},
\label{atom}
\ee
where each $\vD_{\ell_k}$ is an element
of $\R^{\Nxy}$ 
  normalized to unity,
 called `atom'. The $K$-atoms in \eqref{atom} 
  are  selected from a
 redundant set
called a dictionary. 
A {\em{sparse approximation}} of $\vI$ is
an approximation of the form \eqref{atom} such that 
the number of $K$-terms
in the decomposition is significantly smaller
than ${N=N_x N_y}$.

The problem of how to select from a given dictionary 
 the sparsest possible representation of a 
signal is a NP-hard problem \cite{Nat95}. 
In practical applications one looks for 
`tractable sparse' solutions. The mathematical 
methods which are used for this purpose are either 
based on the minimization of the $l_1$-norm 
\cite{CDS01,Eld10} or are greedy strategies
which evolve by stepwise selection of 
atoms from the dictionary \cite{MZ93,PRK93,
RNL02, ARNS04, Tro04, ARN06, DTD06, BD08,  
 NT09, NV10, EKB10}. Greedy strategies are 
 better suited for practical applications. 
In particular, in this work we consider algorithms which 
have been shown effective for approximating 
by partitioning \cite{RNMB13, LRN16}.
\subsection{Approximation of X-Ray medical images 
by partitioning} 
\label{AMI}
A characteristic of X-Ray medical images is that 
they can be best approximated in 
the wavelet domain. This entails to:
\begin{itemize}
\item [(i)] Apply a wavelet transform to the image, i.e.
to convert the intensity image $\vI$ into a 
transformed array $\vW$.
\item [(ii)] Approximate the array $\vW$.
\item [(iii)] Invert the
 approximated  array to recover the
approximated intensity image.
\end{itemize}
The effectiveness of our proposal is based on (a)
the suitability of the proposed dictionary and 
(b) the selection 
approach for approximating the transformed 
image by dividing it into 
 small blocks $\vW_q,\,q=1,\ldots,Q$, 
which we refer to as a `partition' of the array 
$\vW$. 
Without loss of generality 
the blocks are assumed to be square 
of size $\Nq \times \Nq$.  
 We restrict the dictionary to be separable, 
 i.e., 
a $2$D dictionary
$\D=\{\vD_i \in \R^{\Nq \times \Nq}\}_{i=1}^{M}$, 
which is obtained as the tensor product  
$\D=\D^x \otimes \D^y$ of two 1D 
dictionaries 
$\D^x =\{\vd^x_n \in \R^{\Nq}\}_{n=1}^{\Mx}$  and
$\D^y =\{\vd^y_m \in \R^{\Nq}\}_{m=1}^{\My}$, 
with $\Mx \My = M$. 
This represents an important saving in storage. Indeed,  
 instead of having to store a $\Nq^2\times M$ 
 array, only two arrays of 
 size $\Nq\times \Mx $ and $\Nq\times \My$ are to be stored. 
  The reduction in computer memory requirements allows us to 
work with large dictionaries.  

For $q=1,\ldots,Q$ every element $\vW_q$ is approximated 
by an {\em{atomic decomposition}} as below:
\be
\label{atoq}
\vW_q^{k_q}= \sum_{n=1}^{k_q}
c^{k_q,q}(n) \vd^x_{\ell^{x,q}_n}(\vd^y_{\ell^{y_,q}_n})^T,
\ee
where $(\vd^y_{\ell^{y_,q}_n})^T$ indicates the 
transpose of $\vd^y_{\ell^{y_,q}_n} \in \R^{\Nq}$.
The approximated array $\vW^{K}$ is the result of assembling 
the approximated blocks, i.e.,   
$\vW^{K}= \oa_{q=1}^Q \vW_q^{k_q}$, 
where $K=\sum_{q=1}^Q k_q$ and $\oa$ stands for the 
assembling operator, which reconstructs $\vW^K \in \R^{\Nxy}$ from the $Q$ disjoint blocks $\vW_q^{k_q} \in \R^{\Nqq}$. 

The Sparsity Ratio (SR) arising from the approximation is
defined as
$$\SR=\frac{Nx Ny}{K}=\frac{Q \Nq^2}{K}.$$
Our goal is to produce an effective high quality
approximation  with a high value of SR.

The quality of an image approximation is quantified by 
 the Mean Structural SIMilarity (MSSIM)
index~\cite{ssim,ssimex} and the 
 classical Peak Signal-to-Noise Ratio (PSNR), 
 calculated as 
\be
\label{psnr}
\PSNR=10 \log_{10}\left(\frac{{(2^{l}-1)}^2}{\MSE}\right),
\quad 
\MSE=\frac{\|\vI - \vI^K\|_F^2}{\Nx \Ny},\nonumber
\ee
where $l$ is the number of bits used to represent the
intensity of the pixels and
 $\| \cdot \|_F^2$ indicates the Frobenius norm 
induced by the Frobenius inner product, which 
for $\vG_1 \in \R^{\Nx \times \Ny}$ and
$\vG_2 \in \R^{\Nx \times \Ny}$ is defined as 
\be
\la \vG_1, \vG_2 \ra_F = \sum_{i,j=1}^{\Nx, \Ny}
G_1(i,j)  G_2\!(i,j).
\ee
Consequently,
\be
\| \vG_1 \|_F^2=\sum_{i,j=1}^{\Nx, \Ny}
|G_1(i,j)|^2.
\ee 
We aim at achieving high 
PSNR and MSSIM very close to one. 
The question then arises as to how to decide on the
number of atoms, $\kq$, for approximating each block 
in the partition. One possibility is to approximate 
each block totally independently of the 
other blocks and up to 
a fixed tolerance error. This 
possibility has the advantage of enabling straightforward 
parallelization with multiprocessors. Nevertheless, 
linking the approximation of all the blocks through a global 
constraint on sparsity, or quality, 
usually amounts to improving sparsity results 
\cite{RNMB13,LRN16}.
\subsection{Effective greedy strategy for approximating 
by partitioning}
\label{GS}
The common step of the techniques we
consider for constructing 
approximations of the form \eqref{atoq} is the 
stepwise selection of atoms for each block $q$.
On setting $\kq=0$ and  $\vRe_q^{0}=\vW_q$ at 
iteration $\kq+1$ the algorithm selects the indices 
$\ell^{x,q}_{\kq+1}$ and $\ell^{y,q}_{\kq+1}$ 
as follows: 
\be
\label{selec}
\ell^{x,q}_{\kq+1},\ell^{y,q}_{\kq+1}= \operatorname*{arg\,max}_{\substack{n=1,\ldots,\Mx\\
m=1,\ldots,\My}} \left |\la \vd^{x}_n ,\vRe_q^{\kq} \vd^{y}_m \ra_F \right|,
\ee
where $\vRe_q^{\kq}= \vW_q - \vW_q^{\kq}$.  
For the calculation of $\vRe_q^{\kq}$ we find the 
coefficients in \eqref{atoq} through the orthogonal 
projection onto the subspace of selected atoms 
$\V_{\kq}=\Spann\{\vd^{x}_{\ell^{x,q}_n}  
(\vd^{y}_{\ell^{y,q}_n})^T \}_{n=1}^{\kq}$, 
which is equivalent to the minimization 
of $\|\vRe_q^{\kq}\|_F$. For the effective calculation of  
the projections we may choose two different routes, 
depending on the size of the blocks in the partition. 
\begin{itemize}
\item[i)]
Adaptive biorthogonalization of the selected 
atoms $\vA_n^{\kq,q}=\vd^x_{\ell^{x,q}_n} (\vd^y_{\ell^{y,q}_n})^T\in \R^{\Nq\times\Nq}, \,n=1,\ldots,\kq$.
This route gives rise to what is known as the Orthogonal 
Matching Pursuit (OMP) approach \cite{PRK93}. 
Our implementation for separable dictionaries in 2D, 
termed OMP2D \cite{RNBCP12}, involves the 
orthogonalization and re-orthogonalization of the 
selected atoms $\{\vA_n^{\kq,q}\}_{n=1}^{\kq}$, 
producing the orthogonal set 
$\{\vU_n^{\kq,q}, \in \R^{\Nq\times\Nq}\}_{n=1}^{\kq}$, which allows for an effective 
calculation of the set $\{\vB_n^{\kq,q} \in \R^{\Nq\times\Nq}\}_{n=1}^{\kq}$. The 
elements of this set are biorthogonal to the atoms 
$\{\vA_n^{\kq,q}\}_{n=1}^{\kq}$  and are used  to compute the
 coefficients in \eqref{atoq} 
as
\be
\label{coeb}
 c_n^{\kq,q}=\la \vB_n^{\kq,q}, \vI_q \ra_F.
\ee 
\item[ii)]
The Self Projected Matching Pursuit (SPMP) methodology 
\cite{RNB13},
which uses the seminal Matching Pursuit (MP)  method 
\cite{MZ93} as a mean to calculate orthogonal 
projections.
\end{itemize}
The implementation details of the above described OMP2D 
method are given in \cite{RNBCP12} (Appendix A). 
Such an implementation is very effective  
up to some block size. For larger blocks the use of the 
SPMP method, which in 2D for
a separable dictionary
is referred to as SPMP2D \cite{RNB13}, is advised. 
Because it fully exploits 
the separability of dictionaries, the SPMP2D method is 
less demanding in terms of computer memory even though 
 theoretically equivalent to OMP2D.
Full details for its implementation 
are given in \cite{RNB13}. 
The algorithm to realize the self projection 
step is sketched below.

Suppose that for approximating the block $q$  the 
selection process has chosen $\kq$ linearly independent 
atoms labeled by the pair of 
indices $\{ \ell^{x,q}_n , \ell^{y,q}_n \}_{n=1}^{\kq}$ 
and let $\til{\vW}_q^{k_q}$ be an atomic decomposition  
of the form 
\be
\label{atoq2}
\til{\vW}_q^{k_q}= \sum_{n=1}^{k_q}
a^{q}(n) \vd^x_{\ell^{x,q}_n} (\vd^y_{\ell^{y_,q}_n})^T,
\ee
where the coefficients $a^{q}(n), \, n=1,\ldots,\kq$
are arbitrary real numbers.
Every array $\til{\vW}_q \in 
\R^{\Nqq}$ can be expressed as 
\be
\til{\vW}_q= \til{\vW}_q^{k_q} + \til{\vR}.
\ee 
For $\til{\vW}_q^{k_q}$ to be the optimal representation 
of $\til{\vW}_q$ in
$\V_{\kq}= \Spann\{\vd^x_{\ell^{x,q}_n}(\vd^y_{\ell^{y,q}_n})^T\}_{n=1}^{\kq}$, in the sense of minimizing the 
norm of the residue $\vRt$, it should be true that 
$\op_{\V_{\kq}} \vRt=0$. The SPMP2D method fulfills this 
property by approximating $\vRt$ in $\V_{\kq}$, via the 
MP method, and subtracting that component from $\vRt$.  The next algorithm describes the procedure. 
\subsubsection*{Iterative orthogonal projection}
Given a set of previously selected atoms 
$\{\vd^x_{\ell^{x,q}_n}(\vd^y_{\ell^{y,q}_n})^T\}_{n=1}^{\kq}$
set $\vT^0=0$,  $\vRt^{0}=\vRt$,  $j=1$  and 
at each iteration apply the steps below:
\begin{itemize}
\item
Select the pair of indices such that
\be
\label{spmp}
\ell^{x,q}_{j},\ell^{y,q}_{j}= \operatorname*{arg\,max}_{\substack{n=1,\ldots,\kq\\
m=1,\ldots,\kq}} \left | \la \vd^{x}_{\ell^{x,q}_{n}} ,\vRt^{j-1} 
\vd^{y}_{\ell^{y,q}_{m}} \ra_F\right|.
\ee
\item Compute:
\ben
t(j)&=&
\la \vd^x_{\ell^{x,q}_{j}}, \til{\vR}^{j-1}  \vd^y_{\ell^{y,q}_{j}}\ra,\nonumber \\
\til{\vR}^j&=&\til{\vR}^{j-1} - t(j) \vd^x_{\ell^{x,q}_j}  
(\vd^{y}_{\ell^{y,q}_j})^T, \nonumber\\
\vT^j&=& \vT^{j-1} + t(j)\vd^x_{\ell^{x,q}_j}  
(\vd^{y}_{\ell^{y,q}_j})^T.  \nonumber
\een
\item Set $j \leftarrow j+1$ and repeat the process 
until for a given tolerance error $\epsilon$ the condition 
$\| \vT^j- \vT^{j-1}\|_F < \epsilon$ is reached. 
\end{itemize}
The asymptotic exponential convergence of 
 $\vT^j \rightarrow \op_{\V_{\kq}} \vRt$  is proven 
 in \cite{RNS16}.
\subsection{Ranking blocks for the order 
in their approximation} 
\label{hbw}
Especially when an approximation 
 by partitioning is realized 
in the wavelet domain, and the image is sparse in that 
domain, 
it is convenient to impose a global condition on sparsity 
or quality. 
This introduces a hierarchized sequence in which 
the blocks are approximated. The procedure is termed  
 Hierarchized Block Wise (HBW) implementation of
greedy strategies \cite{RNMB13, LRN16}.

Assuming that $\ell^{x,q}_{\kq+1}$ and 
$\ell^{y,q}_{\kq+1},\,q=1,\ldots,Q$ are the 
indices resulting from \eqref{selec}, the block 
to be approximated in the next iteration 
corresponds to the value $q^\star$ such that
$$q^\star= 
\operatorname*{arg\,max}_{q=1,\ldots,Q} 
\left |\la \vd^{x}_{\ell^{x,q}_{\kq+1}}, \vRe_q^{\kq} \,
\vd^{y}_{\ell^{y,q}_{\kq+1}} \ra\right|.
$$
This implies an increment in complexity, with 
respect to the identical strategy without ranking 
the blocks,
of a factor $K$O$(Q)$, with O$(Q)$ accounting for the 
complexity's order for finding the maximum element 
of an array of
length $Q$. As will be illustrated  by the results in 
Table~I, the extra computational cost is in many 
cases compensated by the improvement of sparsity. 
However, the storage requirement of the HBW-OMP2D approach 
 is elevated.  
Notice that, in the implementation of OMP2D discussed 
above, the HBW version needs to store at least 
the orthogonal sets 
$\{\vU_n^{\kq,q}, \in \R^{\Nq\times\Nq}\}_{n=1}^{\kq}$ 
for each of the blocks in the partition, only for the
 realization of the orthogonal projection. An implementation 
of the same strategy, 
requiring much less computer memory, realizes the orthogonal 
projection via the above iterative projection algorithm.
Such an approach is the HBW version of the SPMP2D method 
(HBW-SPMP2D) \cite{RNS16}.
\subsection{HBW Pruning}
\label{hbwb}
In this section we consider the downgrading of a given 
approximation when carried out in a HBW  manner. 
To this end, firstly the approximation of each block  
is realized,
up to the same tolerance error, 
totally independently of the other blocks. This 
leaves room for the possibility of   
parallelization of the block approximation 
when multiprocessors are available. The outcome of 
this stage is an approximation of the form 
\eqref{atoq} for every block $q=1,\ldots,Q$. The 
second stage consists in slightly downgrading the 
approximation by pruning some of the coefficients 
in the atomic decomposition of the blocks, in a 
HBW fashion. As mentioned  in  the previous section, 
 the HBW version of a greedy strategy for 
approximating by partitioning ranks the blocks for 
their sequential approximation.
 The optimized way of downgrading an approximation in a 
HBW manner is termed HBW  Backwards Optimized Orthogonal Matching Pursuit   
(HBW-BOOMP) \cite{LRN16}. For large images this
 approach is demanding 
in terms of storage. A method with less 
memory requirements, though not optimized, 
is termed HBW Backwards Self Projected Matching pursuit
(HBW-BSPMP) \cite{RNS16}. This  method 
downgrades an 
 approximation by disregarding
some coefficients in a HBW fashion. 
Given the approximation of a partition as in 
\eqref{atoq}, 
the HBW-BSPMP algorithm in 2D (HBW-BSPMP2D) 
iterates as follows: 
\begin{itemize}
\item[1)] For $q=1,\ldots,Q$
select the `potential' coefficient $c^{q}(j^q)$
to be eliminated from the atomic decomposition of every
block $q$, according to the criterion:
$$j^{q}=
\operatorname*{arg\,min}_{n=1,\ldots,\kq} |c^q(n)|^2.$$
\item[2)]Select the block $\qd$ such that
$$\qd = \operatorname*{arg\,min}_{q=1,\ldots,Q}
 |c^q(j^q)|^2$$
and downgrade the atomic decomposition of the block $\qd$
 by removing the atoms corresponding to the
indices $\ell^{x,\qd}_{\jqd}, \ell^{y,\qd}_{\jqd}$.  
 This produces the additional residual component
 $ \Delta \til{\vR}_{\qd}= c^{\qd}\!(\jqd) 
 \vd^x_{\ell^{x,\qd}_{\jqd}} (\vd^y_{\ell^{y,\qd}_{\jqd}})^T   $. 
\item[3)] Approximate  $ \Delta \til{\vR}_{\qd}$
 in $\Spann\{\vd^x_{\ell^{x,\qd}_{n}} (\vd^x_{\ell^{x,\qd}_{n}})^T\}_{\!\!\!\!{\substack{\!\!n=1\\\,\,\,\,\, n \ne  \jqd}}}^{\kqd}$
 using the projection algorithm 
given in Sec.~\ref{GS} to obtain
$$ \Delta \til{\vR}_{\qd}=
\sum_{i=1}^{\kq} t(i) \vd^x_{\ell^{x,\qd}_{n}} (\vd^y_{\ell^{y,\qd}_{n}})^T.$$ 
Update the coefficients in the atomic decomposition 
of the block $\qd$ as
$$\left\{c^{\qd}\!(n)\right\}_{\!\!\!\!{\substack{\!\!n=1\\ 
\,\,\,\,\, n \ne \jqd}}}^{\kqd} 
\leftarrow \left \{ c^{\qd}\!(n) + t(n) \right\}_{\!\!\!\!{\substack{\!\!n=1\\\,\,\,\,\, n \ne  \jqd}}}^{\kqd}.$$
\item[4)] 
Shift the indices of the coefficients
 and atoms in the decomposition corresponding to the
 block $\qd$, to allow for the removal of 
the  $\jqd$-th term. 
\item[5)]
Set $\kqd \leftarrow \kqd-1$ and
check if  the stopping criterion has been met.\\
Otherwise:
\begin{itemize}
\item[$\bullet$]
Select a new potential coefficient to be removed from
the atomic decomposition of block $\qd$ using the
same criterion as in 1).
\item[$\bullet$] Repeat steps 2) - 5).
\end{itemize}
\end{itemize}
\subsection{Constructing suitable dictionaries for X-Ray
medical images}
The degree of success in achieving high sparsity
using a dictionary approach, depends on the suitability
of the dictionary. One possibility to produce a `good'
dictionary is to learn it from training data. In the
last decade a number of techniques for learning
dictionaries have been proposed \cite{KMR03, AEB06, RZE10,
TF11, ZGK11, SNS15}. 
Most techniques, though, are not designed 
for learning large separable dictionaries. 
 In this work we propose a
separable dictionary, which is very easy to construct
and allows us to achieve the goals of the paper.
It is a mixed dictionary
consisting of two classes of sub-dictionaries of
different nature:
 I) The trigonometric dictionaries  $\mathcal{D}_{C}^x$ and
$\mathcal{D}_{S}^x$, defined below,
\ben
\mathcal{D}_{C}^x\!\!&=&\!\!\{w_c(n)
\cos{\frac{{\pi(2i-1)(n-1)}}{2M}},i=1,\ldots,N\}_{n=1}^{M}\nonumber\\
\mathcal{D}_{S}^x\!\!&=&\!\!\{w_s(n)\sin{\frac{{\pi(2i-1)(n)}}{2M}},i=1,\ldots,N\}_{n=1}^{M},\nonumber
\een
where $w_c(n)$ and $w_s(n),\, n=1,\ldots,M$
are normalization factors.
II) The dictionary $\mathcal{D}_{L}^x$,
which is constructed
by translation of the prototype atoms in Fig.1. Notice that 
those prototypes are `Hadamard-like' atoms, but with
 support one, two, and three.
The mixed dictionary $\mathcal{D}^x$
is built as
$\mathcal{D}^x = \mathcal{D}_{C}^x \cup \mathcal{D}_{S}^x
\cup \mathcal{D}_{L}^x$ and
$\mathcal{D}^y= \mathcal{D}^x$. The  concomitant
 2D dictionary, $\D= \mathcal{D}^x \otimes \mathcal{D}^y$,
 may be very large, but never needed as such.
\begin{figure}[h!]
\begin{center}
\includegraphics[width=9cm]{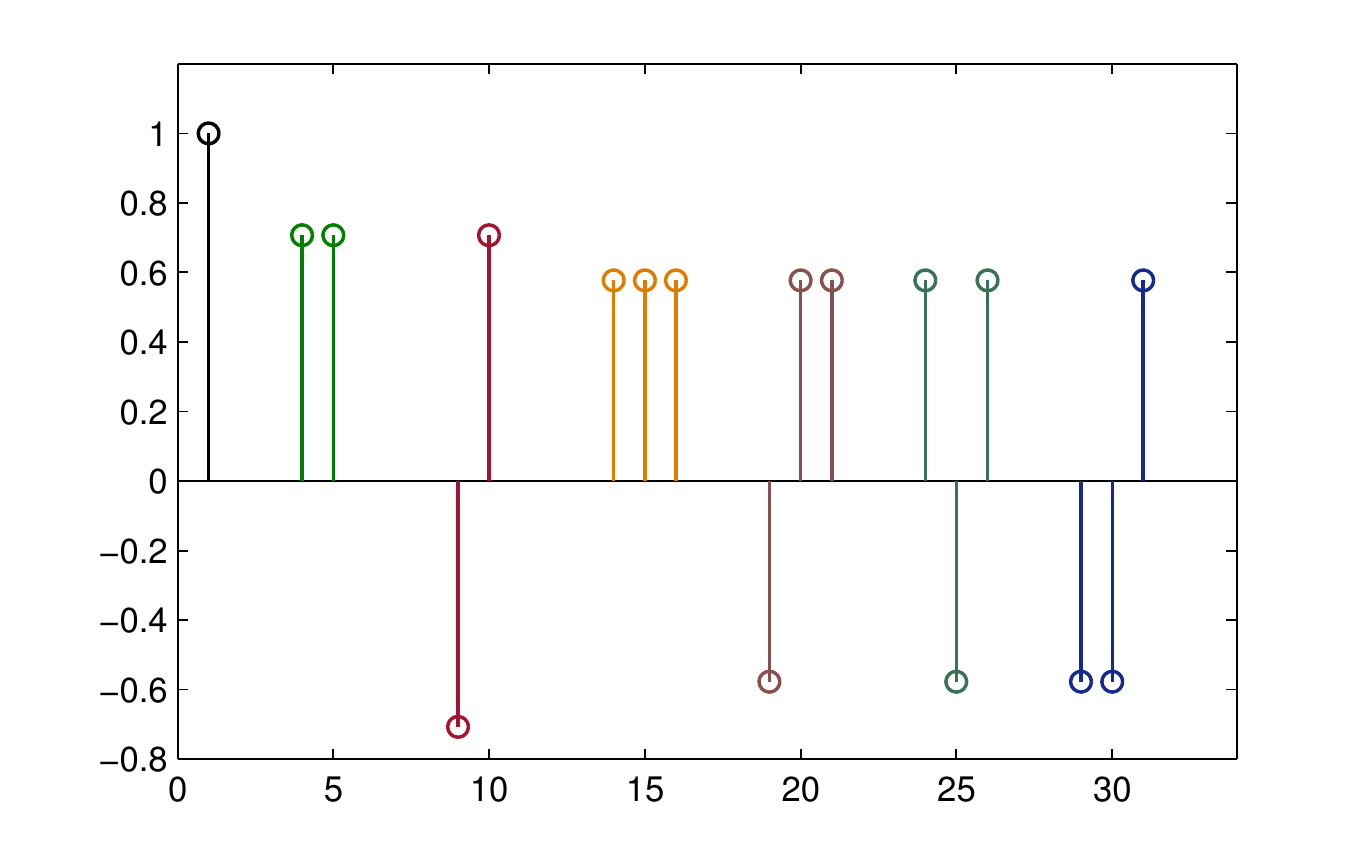}
\end{center}
\vspace{-0.5cm}
\caption{{\small{Prototype atoms, which generate the
dictionaries $\mathcal{D}_{L}^x$ by sequential
translations of one point. Each prototype is shown in a 
different color.}}}
\end{figure}
The graphs in Fig.2 depict four different
2D atoms.
\begin{figure}[h!]
\begin{center}
\includegraphics[width=9cm]{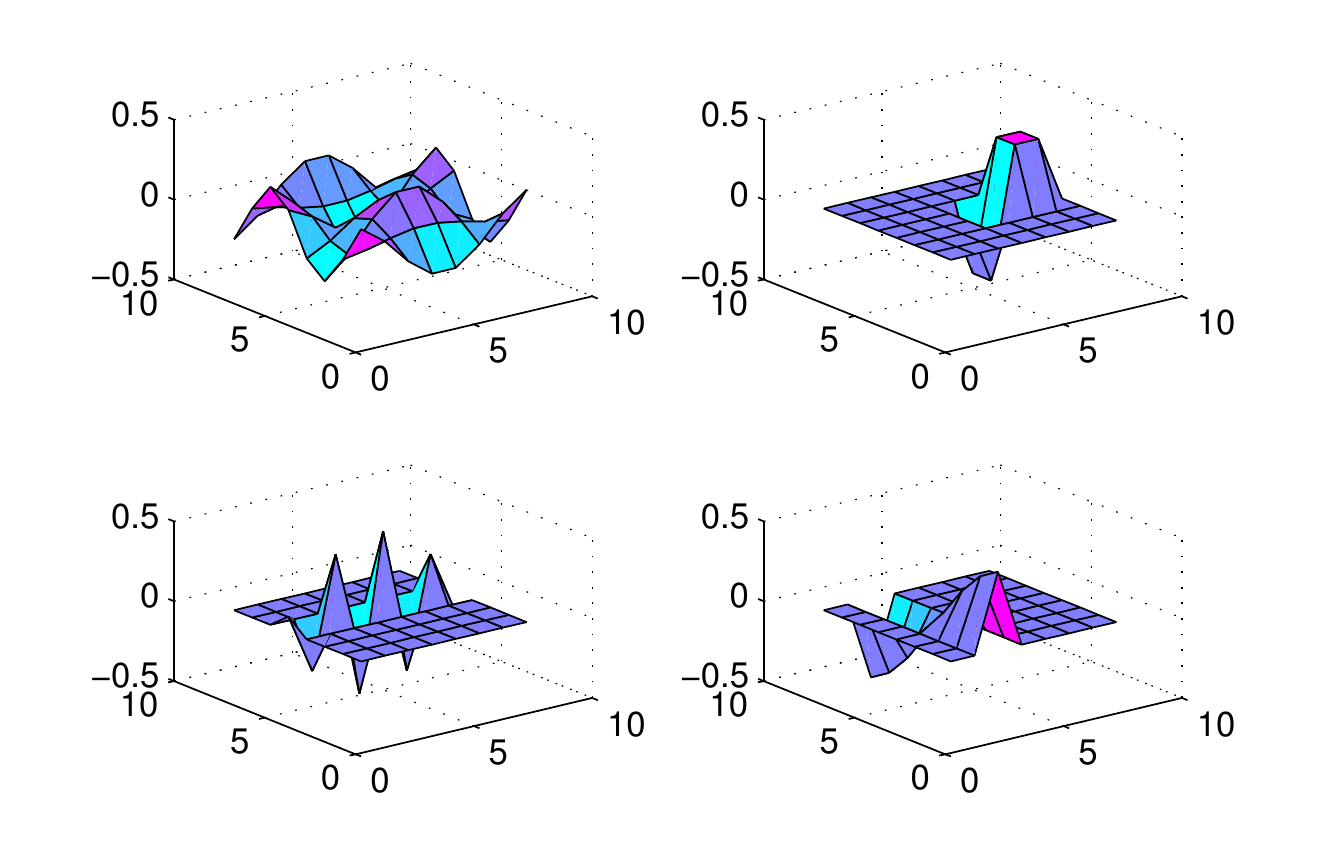}
\end{center}
\vspace{-0.5cm}
\caption{{\small{Four different 2D atoms,  
$\vd^x_{\ell_i} (\vd^y_{\ell_j})^T$, with
$\vd^x_{\ell_i}$ a member of $\mathcal{D}_{C}^x$
  and $\vd^y_{\ell_j}$ a member of $\mathcal{D}_{C}^y$
(top left graph), $\vd^x_{\ell_i}$ a member of   $\mathcal{D}_{L}^x$ and $\vd^y_{\ell_j}$
a member of  $\mathcal{D}_{L}^y$ (top right graph), 
 $\vd^x_{\ell_i}$ a member of   $\mathcal{D}_{L}^x$ and $\vd^y_{\ell_j}$
a member of  $\mathcal{D}_{S}^y$ (bottom left graph),
 $\vd^x_{\ell_i}$ a member of   $\mathcal{D}_{C}^x$ and $\vd^y_{\ell_j}$
a member of  $\mathcal{D}_{L}^y$ (bottom right graph)}}}
\end{figure}\\

{\bf{Remark 1:}} The mixed  dictionary given above
 is specially suited for 
approximations in the wavelet domain. 
Once the approximation of the  blocks is concluded,
these are assembled to produce the approximated array
 $\vW^{K}= \oa_{q=1}^Q \vW_q^{k_q}$. Finally,
the inverse wavelet transform
is applied to convert the array $\vW^{K}$ into the
approximation of the intensity image $\vI^K$.
\subsection*{Numerical Examples}
We illustrate now the suitability of the proposed mixed 
dictionary to produce 
high quality approximations of the set of X-ray 
medical images shown in Fig.~\ref{images}. 
This set of twenty images is the Lukas 2D 8 bit medical image corpus, available on \cite{lukas}.
\begin{figure}[h!]
\label{images}
\begin{center}
\vspace{-2cm}
\hspace{-1.5cm}
\includegraphics[width=4.9cm]{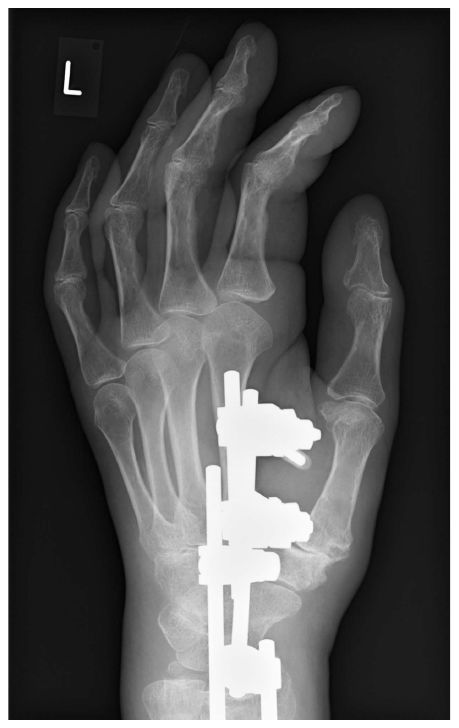}\hspace{-3.5cm}
\includegraphics[width=4.9cm]{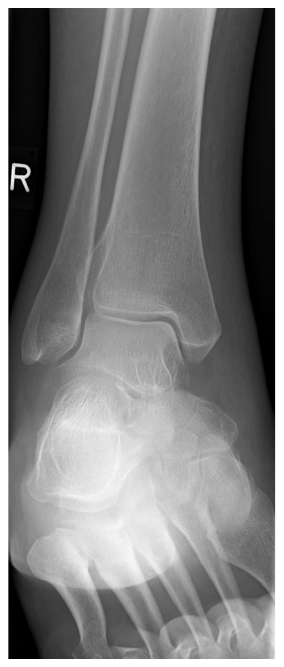}\hspace{-3.5cm}
\includegraphics[width=4.9cm]{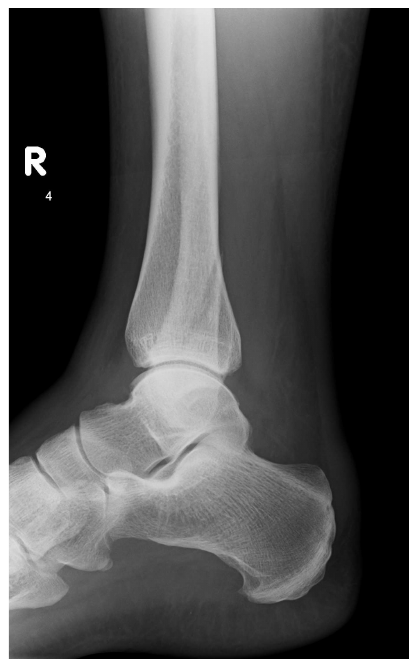}\hspace{-3.5cm}
\includegraphics[width=4.9cm]{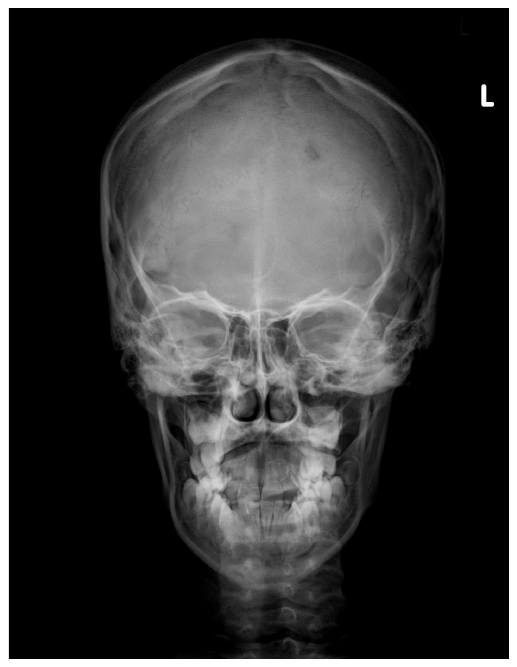}\\
\vspace{-5cm}
\hspace{-1.5cm}
\includegraphics[width=4.9cm]{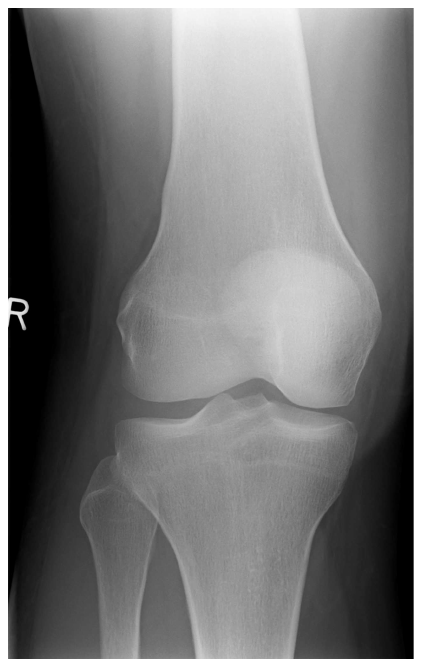}\hspace{-3.5cm}
\includegraphics[width=4.9cm]{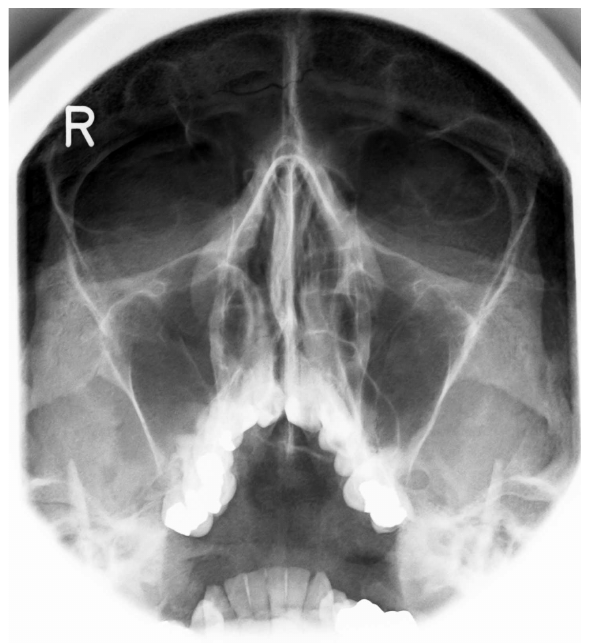}\hspace{-3.5cm}
\includegraphics[width=4.9cm]{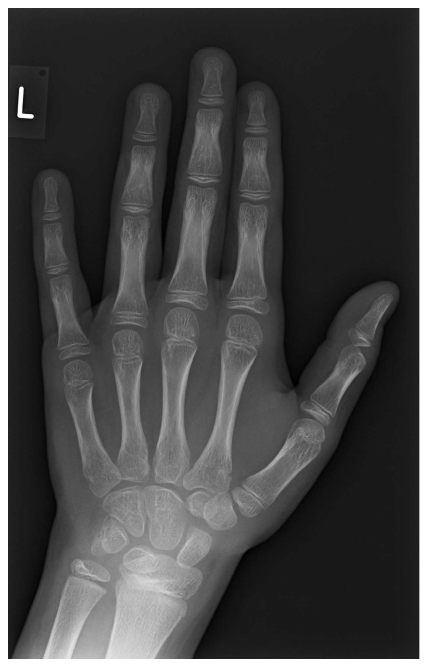}\hspace{-3.5cm}
\includegraphics[width=4.9cm]{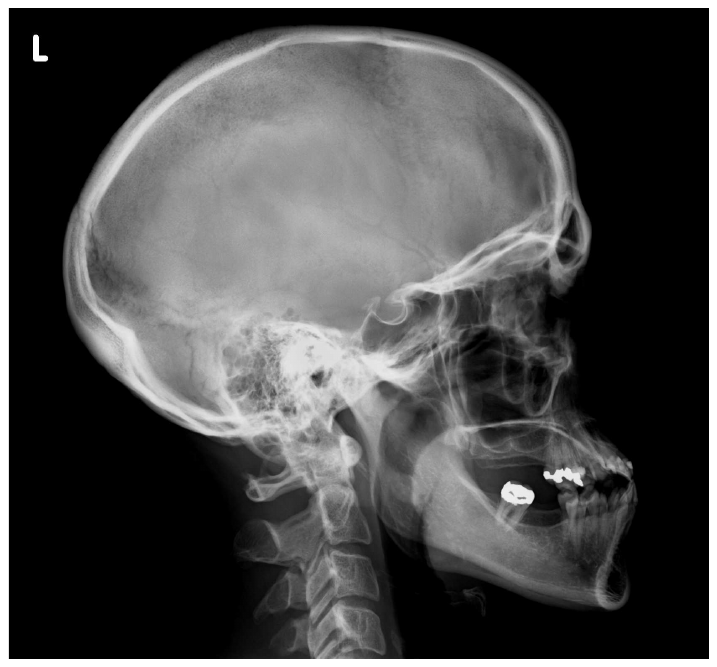}\\
\vspace{-5cm}
\hspace{-1.5cm}
\includegraphics[width=4.9cm]{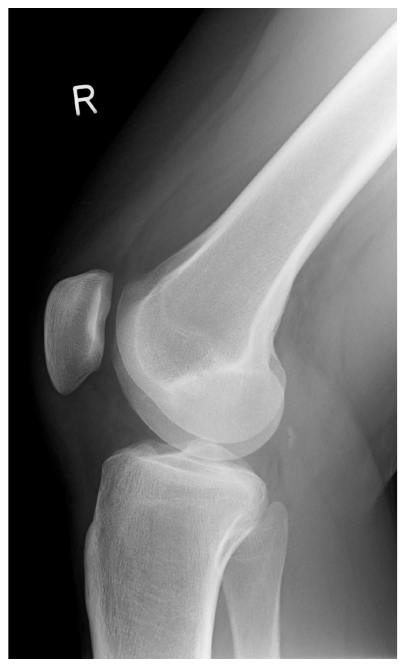}\hspace{-3.5cm}
\includegraphics[width=4.9cm]{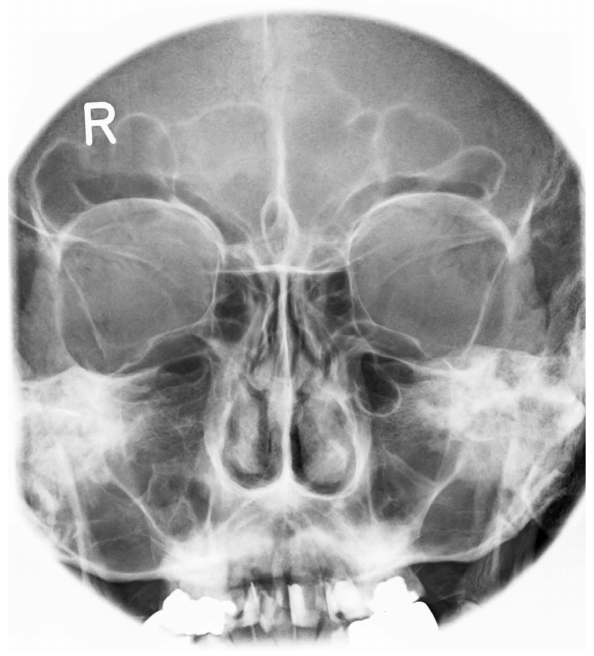}\hspace{-3.5cm}
\includegraphics[width=4.9cm]{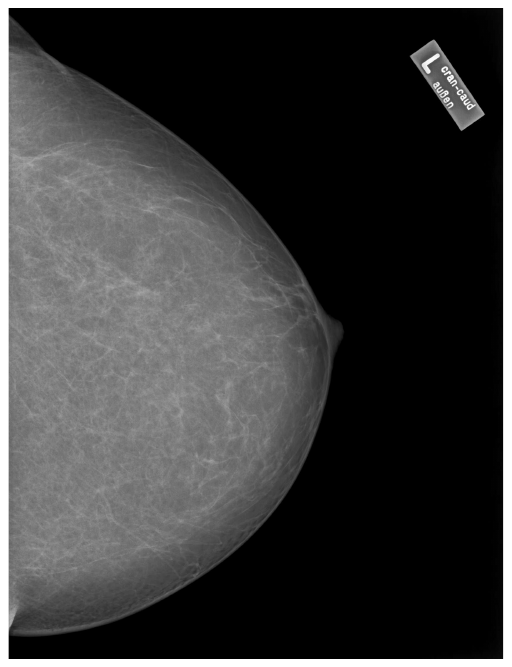}\hspace{-3.5cm}
\includegraphics[width=4.9cm]{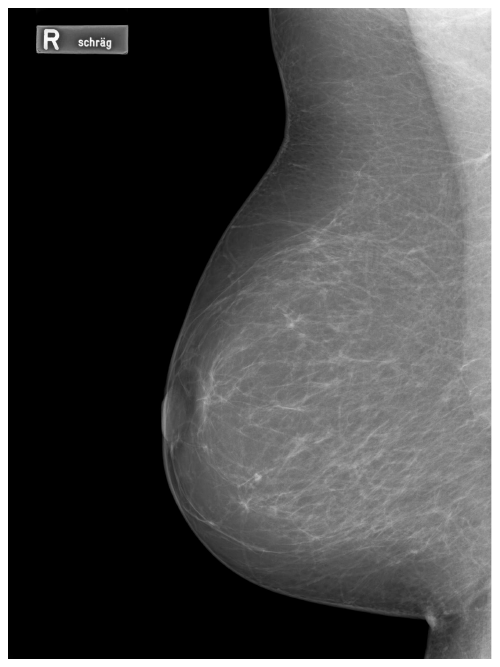}\\
\vspace{-4.7cm}
\hspace{-1cm}
\includegraphics[width=4.3cm]{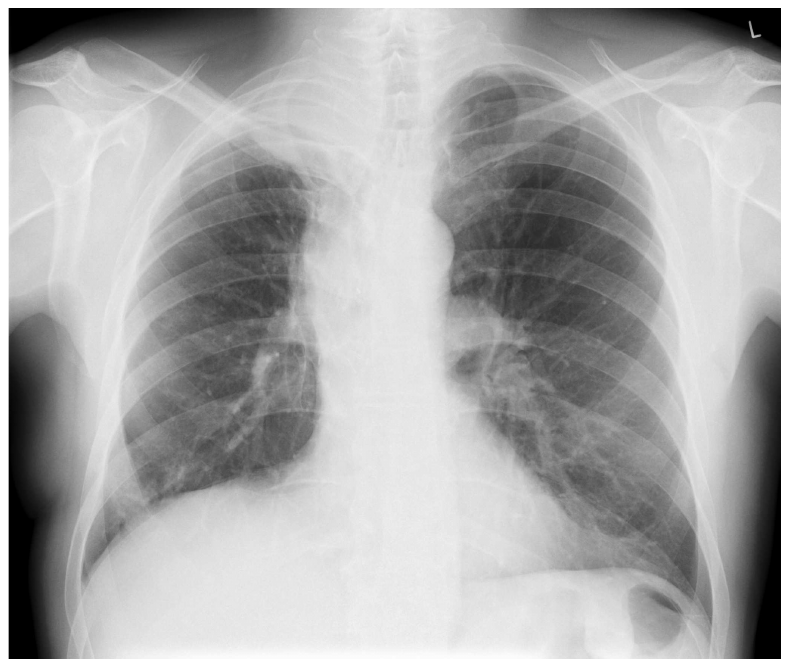}\hspace{-2.6cm}
\includegraphics[width=4.3cm]{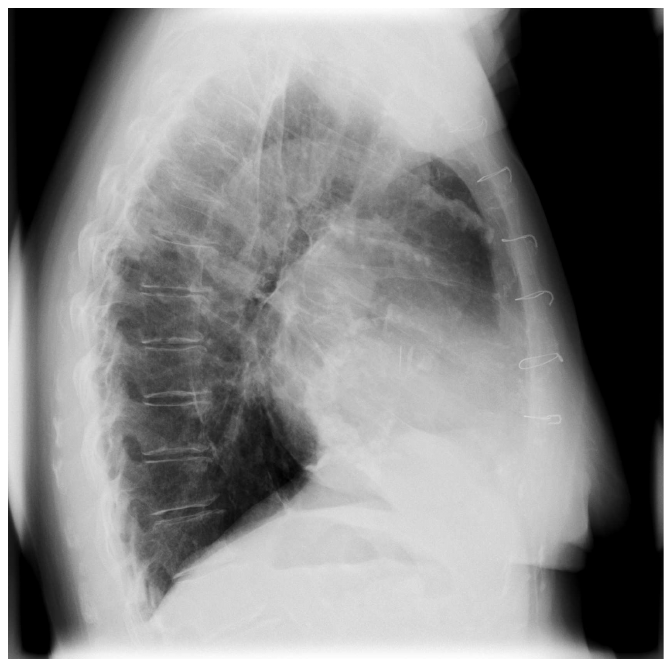}\hspace{-2.6cm}
\includegraphics[width=4.3cm]{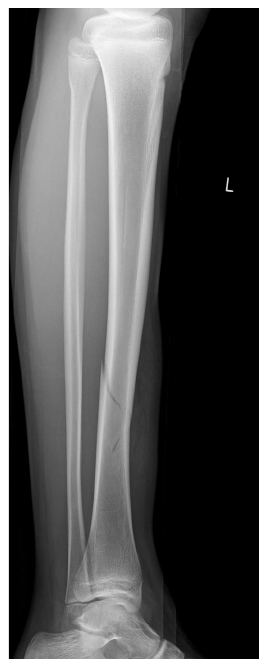}\hspace{-3.5cm}
\includegraphics[width=4.3cm]{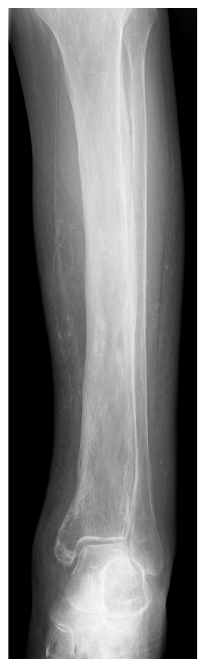}\\
\vspace{-4.3cm}
\hspace{-1cm}
\includegraphics[width=4.3cm]{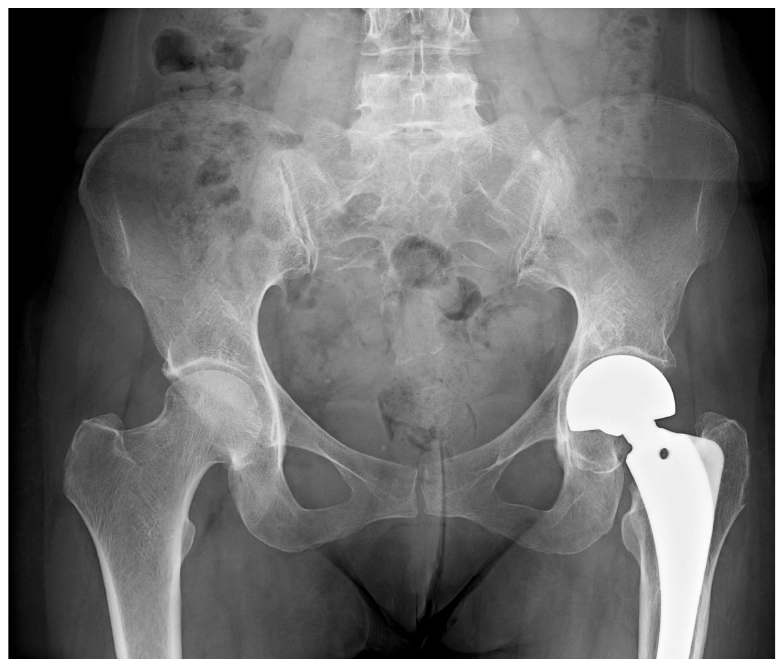}\hspace{-2.6cm}
\includegraphics[width=4.3cm]{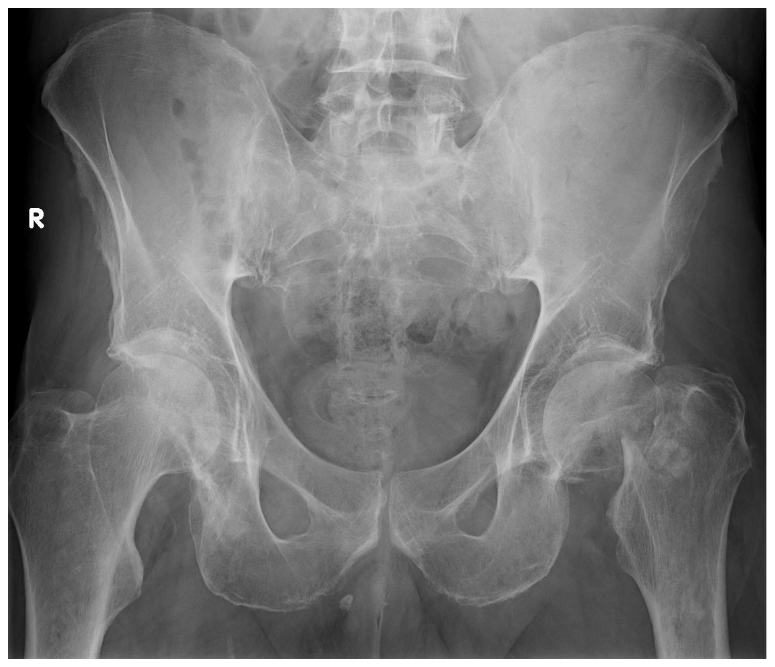}\hspace{-2.6cm}
\includegraphics[width=4.3cm]{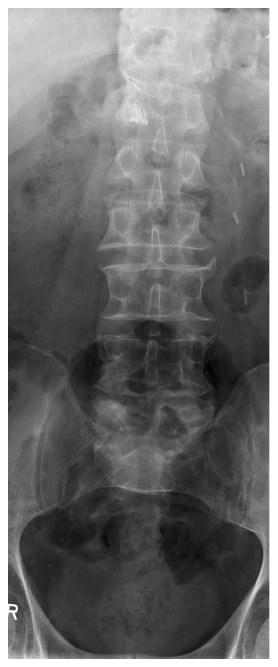}\hspace{-3.5cm}
\includegraphics[width=4.3cm]{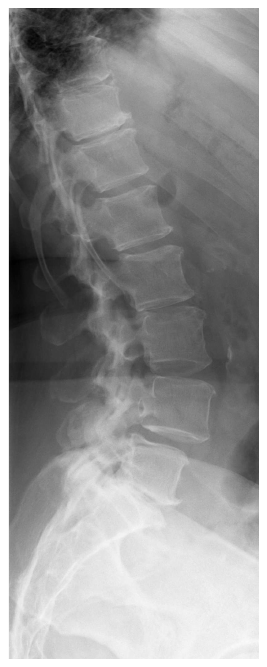}
\end{center}
\vspace{-2.1cm}
\caption{\small{Lukas Corpus  \cite{lukas} listed in Table~I. From top to bottom and left to right: Hand$_0$, Foot$_1$, Foot$_0$, 
Head$_0$, Knee$_1$, 
 Sinus$_0$, Hand$_0$, Head$_1$, Knee$_0$, Sinus$_1$, 
 Breast$_0$, Breast$_1$,  Thorax$_0$, Thorax$_1$, Leg$_0$. 
 Leg$_1$, Pelvis$_1$, Pelvis$_0$, Spine$_1$, 
Spine$_0$.}}
\label{images}
\end{figure}\\
The comparison is performed with respect to the SR 
achieved by nonlinear thresholding of 
the DCT and DWT coefficients, respectively, to 
produce an approximation of the same quality. 
The quality is set by requiring a value of 
MSSIM greater than 0.998, and fixing the 
PSNR as a value for which the DWT approach produces
the required MSSIM. The PSNR is more sensitive 
to small variations in the approximation than the MSSIM.
By fixing the PSNR, at the values given in 
Table~I, the equivalence of quality is ensured
with respect to both measures. 
Certainly, in its original size is not possible 
to distinguish the actual images from their 
approximations with any of the approaches.  

The first column of Table~I lists the images in 
Fig.~\ref{images}. The second column displays 
the value of PSNR. The third and 
forth columns correspond to the SR obtained through the 
DWT and DCT, respectively.      
The DWT approximation is calculated by applying the
Cohen-Daubechies-Feauveau 9$/$7 (CDF97) wavelet transform 
on the whole image (using the {\tt{waveletcdf97}} MATLAB  routine available on \cite{CDF97})  
 and reducing coefficients, by 
iteratively thresholding, to produced the 
 required quality in the pixel-intensity domain.
The DCT approximation is calculated by applying the 
 DCT to the intensity image 
 on a partition of block size $8 \times 8$
and reducing coefficients in a HBW manner. This 
procedure and block size yield  
the best sparsity results using DCT.
The images are ordered according to 
the SR achieved with the DWT approximation. The 
sparsest is the first image and the least sparse 
the last one.

The first step in the implementation of 
the dictionary approximation is to transform the image,
for which  we use the {\tt{waveletcdf97}} function.
 For the images in the upper part of the 
table (from Head$_1$ to Breast$_0$) the SR results 
 rendered by this approach, for a partition of 
block size $8 \times 8$, strongly depend on the method 
used for the selection process.  Because those images are 
sparse in the wavelet domain, the ranking of 
the blocks for their approximation through the HBW-OMP2D 
method yields significantly higher SR (sixth column)
 than the standard application of OMP2D (fifth column). 
\begin{table}[!h]
\begin{center}
\begin{adjustbox}{max width=\textwidth}
\begin{tabular}{|l|c||r|r|r|r|}
\hline
Image & PSNR & DWT & DCT & OMP2D& HBW 
 \\ \hline \hline
Hand$_1$& 48.1& 30.0&
26.4&  39.0  & 72.6\\ \hline
Foot$_1$ &48.6& 26.6&
26.1&30.4  & 44.9 \\ \hline
Foot$_0$&48.6&25.5&
26.1&  42.7  & 65.2 \\ \hline
Head$_0$& 47.4& 25.3&
24.3& 51.9&    63.2 \\ \hline
Knee$_1$& 48.0 & 22.7&
23.0&  34.5 &   59.8 \\ \hline
Sinus$_0$&47.1 &18.9&
18.7& 31.3 &  46.7 \\ \hline
Hand$_0$& 48.8&18.6&
18.7& 32.2& 47.9 \\ \hline
Head$_1$&46.4&17.5&
15.1& 38.3  & 44.4 \\ \hline
Knee$_0$& 49.1 & 17.4&
17.5&  33.2 &  45.9 \\ \hline
Sinus$_1$& 45.8 &17.2&
17.1& 29.5 &  43.0 \\ \hline
Breast$_0$&44.3&15.7&
15.3& 36.7 & 41.0 \\ \hline \hline
Breast$_1$& 44.3& 11.5&
11.2& 27.7& 29.7 \\ \hline
Thorax$_0$&44.1&10.6&
10.9& 25.1&  27.4 \\ \hline
Thorax$_1$&43.4&10.3&
9.6& 25.4& 26.3 \\ \hline
Leg$_0$&48.9&8.2&
8.4& 21.2& 22.3 \\ \hline
Leg$_1$&49.2&5.8&
5.9& 15.1& 15.4 \\ \hline
Pelvis$_1$&44.3&4.8&
4.7& 12.3&  12.6 \\ \hline
Pelvis$_0$&44.4&4.6&
4.7& 12.4& 12.6\\ \hline
Spine$_1$&47.0&3.5&
3.6&9.3& 9.4 \\ \hline
Spine$_0$&47.4&2.9&
2.8 & 7.1&7.7 \\ \hline
\end{tabular}
\end{adjustbox}
\end{center}
\caption{Comparison of the SR calculated by the
DWT, DCT, and dictionary approaches. The
approximation with DCT and dictionary are carried out
on a partition of  
 $8\times8$ blocks. The values corresponding to the
dictionary are those obtained with the
OMP2D method and its corresponding HBW version.
All the approximations produce a MSSIM  
greater than 0.998.}
\end{table}\\
The images in the lower part of
Table~I (from  Breast$_1$ to Spine$_0$) 
 are less sparse in the wavelet domain, 
as indicated by the SR produced by the DWT approximation. 
Thus, even if the dictionary method achieves, for all the images,  a  
very significant gain in SR  with respect to the DWT and DCT 
approaches, the results corresponding to the
OMP2D and HBW-OMP2D methods are much closer
than they are for the images in the upper part of the table. 
As can be seen in Table II, the processing time with both
methods is very competitive, considering that
the results have been produced in a MATLAB environment
using C++ MEX files for the OMP2D and HBW-OMP2D routines.
\begin{table}[!h]
\begin{center}
\begin{adjustbox}{max width=\textwidth}
\begin{tabular}{|l|c|c|c|}
\hline
Corpus &Mean value size& OMP2D & HBW 
\\\hline \hline
X-Ray Lukas & $1943\times 1364$ pixels & 5.1 secs &
10.3 secs\\ \hline 
\end{tabular}
\end{adjustbox}
\end{center}
{\caption{Mean value processing time (per image) to obtain the results of Table I. The second column shows the  mean value size of the images in the set.  The third column corresponds to the OMP2D method and the forth column to the HWB-OMP2D one.  The given times are the  average of five independent runs with a single processor in a notebook Core i7 3520M, 4GB RAM.}}
\end{table}

While for the images in the upper part of the table
is worth applying
the HBW-OMP2D method, for most
of the images in the lower part of the table the
difference is not significant.
All further results of SR will be presented by 
grouping the first eleven images (from Hand$_1$ to 
Breast$_0$) in a set, say $\cal{U}$, 
and the remaining images (from  Breast$_1$ to Spine$_0$) 
 in a set $\cal{L}$. 

For the same quality as in Table~I , with the dictionary 
approach we also consider partitions 
of block size $16 \times 16$ and 
$24 \times 24$. The points in the top graph of Fig.~3 
 represent the mean vale of the SR 
with respect to the images in the set $\cal{U}$, 
 vs  block size $8 \times 8 , 16 \times 16$, 
and $24 \times 24$.
In order to facilitate a visual comparison, the DCT and DWT 
results are simply repeated.  They correspond to
the best result for the DWT, which occurs when
each image is processed as a whole, and the best result
for DCT, which occurs for a partition of block size $8 \times 8$.
\begin{figure}[h!]
\begin{center}
\includegraphics[width=9cm]{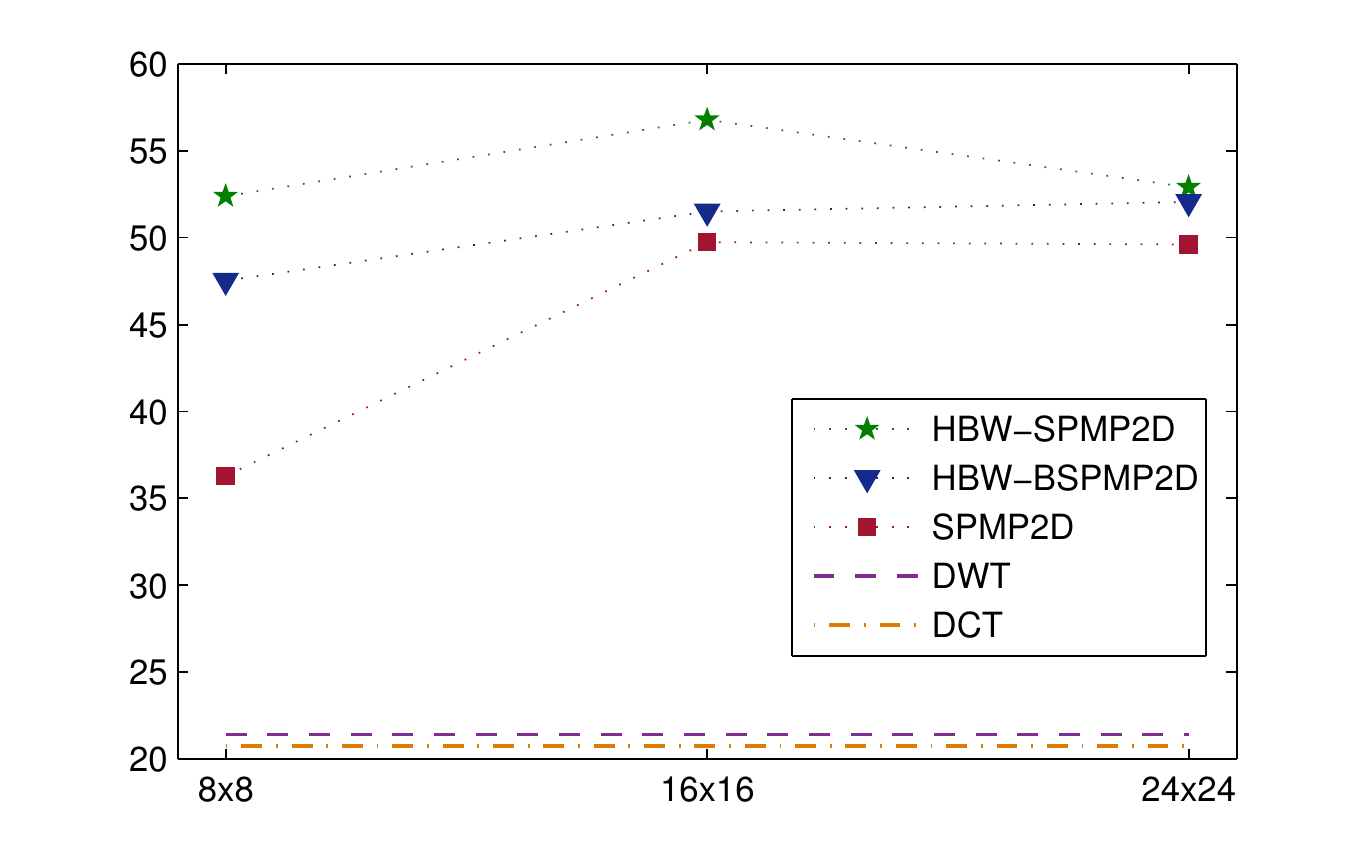}\\
\includegraphics[width=9cm]{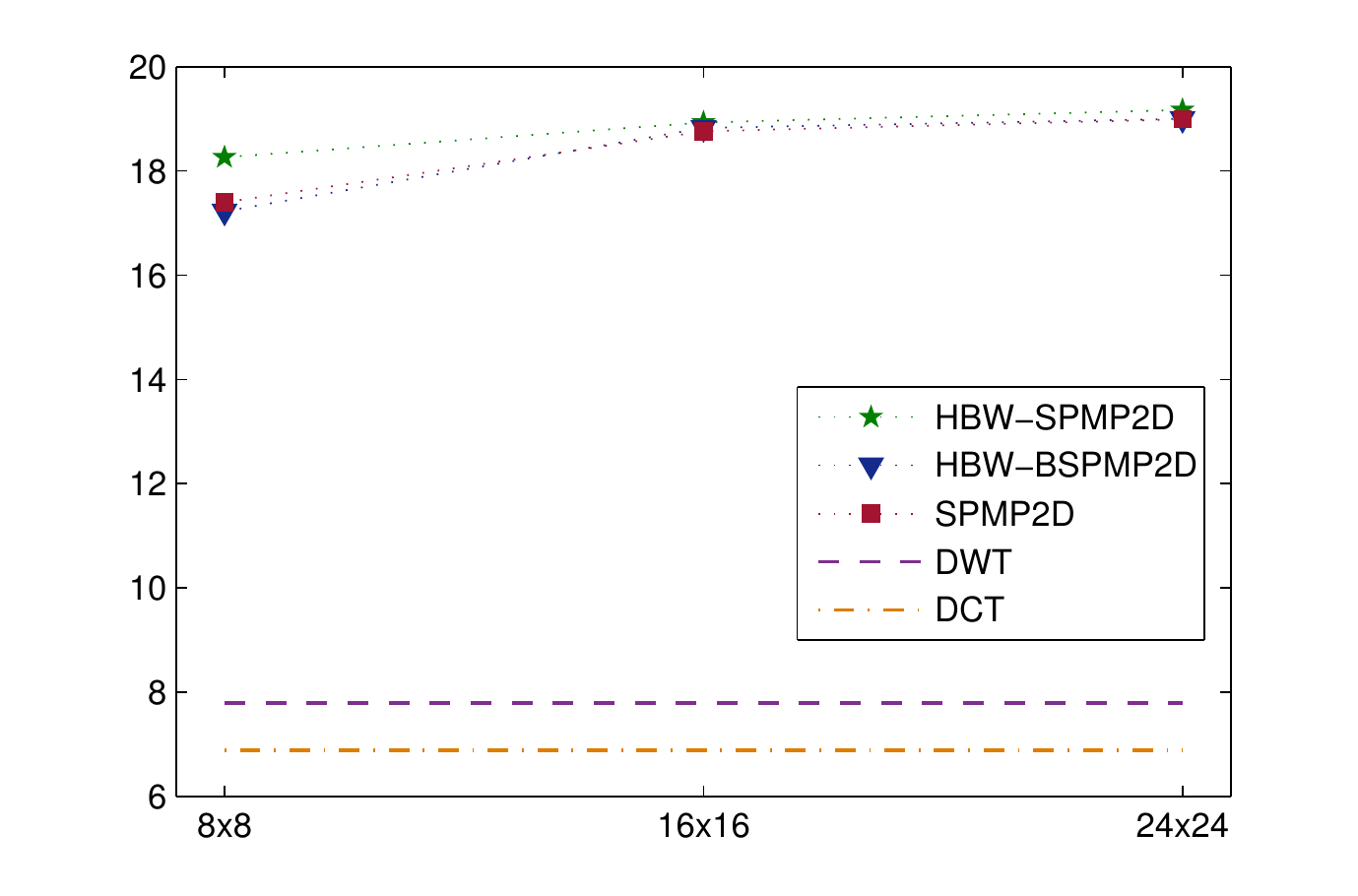}
\vspace{-0.3cm}
\caption{\small{Mean SR vs block size  
$\Nq \times \Nq$  
for the X-Ray images in Fig.~\ref{images}. 
The green stars are the values obtained with HBW-SPMP2D, 
the red squares with SPMP2D, and the 
blue triangles with  HBW-BSPMP2D. 
The dashed line and the 
dash-dot line represent the DWT and DCT results, respectively.
The top graph  corresponds to the images in the 
upper part of Table~I (set ${\cal{U}}$) and the bottom one 
 corresponds to the images in the lower 
part of that table (set ${\cal{L}.}$)}}
\label{SR2}
\end{center}
\end{figure}

Due to memory requirements, for the images in the  set
${\cal{L}}$  and for the partition of 
 block of size  $24 \times 24$ we can only implement
HBW-SPMP2D. 
Hence, for consistency, the graphs 
in Fig.~4 have been produced with the
SPMP2D and HBW-SPMP2D methods. The red squares  
correspond to the mean SR value  obtained by SPMP2D and the 
green stars are those obtained by HBW-SPMP2D. 
The blue triangles in the same graphs  are the result of 
 applying a combination 
of approaches. Firstly, each image is approximated 
block by block with SPMP2D up to a PSNR $2\%$ higher 
than the required 
PSNR. Subsequently, the approximation is downgraded in a 
HBW fashion to produce the required global PSNR by means of
 the HBW-BSPMP2D approach, as described in 
Sec.~\ref{hbwb}.
The top graph in Fig.~4 corresponds to  
the images in the set ${\cal{U}}$, and the difference 
between approaches is noticeable. On the contrary, 
as seen in the bottom graph  
corresponding 
to the images in the set ${\cal{L}}$, except for 
 block size $8 \times 8$ all the 
three methods yield equivalent results.\\

{\bf{Remark 2:}} The numerical tests presented in this 
section lead to the following conclusions:
\begin{itemize}
\item
Approximating the set of medical images in Fig.~\ref{images} using the proposed mixed dictionary produces a very 
significant gain in the mean $\SR$  value, 
in relation to the one yielded by 
traditional DCT and DWT  approximations. 
\item
The best compromise between sparsity and computational time 
is attained for a partition of block size $16 \times 16$.
The quantification of the relative gain in SR
of one particular approach, in relation to other,
is given by the quantity:
$$\G= \frac{\SR_A - \SR_B}{\SR_B} 100 \%,$$
where $\SR_B$ is the SR produced by the
 approach for which the gain is referred to. 
Accordingly, 
the mean value sparsity gain  
in relation to DWT results, including  all the twenty 
images in the set, 
is  $142\%$ with standard 
deviation of $18\%$. These results can be obtained very 
effectively through the HBW-OMP2D 
method, if the images are sparse in the wavelet domain. 
Otherwise, the OMP2D method is more effective 
because it produces faster equivalent results. 
Moreover, parallelization of block approximation 
with multi-processors is straightforward.
\item
For blocks larger than $16 \times 16$ the
methods SPMP2D and HBW-SPMP2D (low memory 
implementations of OMP2D and HBW-OMP2D respectively) 
may be required. However, when the approximation is 
carried out in the wavelet domain blocks of size larger than 
$16 \times 16$ do not improve sparsity in a significant 
way.
\item
For partitions of block size $8 \times 8$, 
refining a OMP2D approximation by HBW-BSPMP2D pruning is
an option worth consideration, if the image is 
sparse in the wavelet domain. The actual results vary 
according to how far the forward selection goes. 
The results presented here correspond to a slight 
pruning which degrades the quality of 
the forward approximation only $2\%$.
\item
Since the DWT is a fast approximation, it can be
used as a tool to help decide the strategy for 
approximating with dictionaries. If the fast DWT 
approximation gives a high $\SR$ (say $\SR> 10$ for the 
high quality  reconstruction required in this context) 
then using the HBW version of a pursuit strategy is 
strongly advised. 
\end{itemize} 

{\bf{Note:}} All the routines for implementation
 of the approximation methods 
 and the script for reproducing results 
have been made available on the website \cite{NLAweb}.
\section{Conclusions}
Sparse representation of X-Ray medical images in the 
context of data reduction 
has been considered. The success of the framework is 
based on (a) the suitability of the proposed
dictionary and (b) the effectiveness of the  
algorithms for realizing the approximation. The 
comparison with traditional approaches such as 
non linear approximations through DWT and DCT 
redounds in a mean value gain of 
sparsity of up to $148\%$ (for block size $16 \times 16$) 
which is 
achieved at a very competitive time (11.4 secs per image)
even when 
the implementation is carried out  
in a small notebook within a MATLAB 
environment. The results are really encouraging. 
We feel confident that the proposed framework 
will be of assistance to X-ray image processing 
applications relying on data reduction. In order 
to facilitate further developments, as well as the 
reproduction of the results in this paper, 
the implementation of all the algorithms 
has been made available on 
a dedicated website \cite{NLAweb}.  

\end{document}